%% file: paper.tex
\documentclass[10pt,twocolumn,letterpaper]{article}

\usepackage{cvpr}
\usepackage{times}
\usepackage{epsfig}
\usepackage{graphicx}
\usepackage{amsmath}
\usepackage{amssymb}
\usepackage{multirow}
\usepackage{float}

\usepackage{listings}
\usepackage{color}

\definecolor{dkgreen}{rgb}{0,0.6,0}
\definecolor{gray}{rgb}{0.5,0.5,0.5}
\definecolor{mauve}{rgb}{0.58,0,0.82}

\usepackage[pagebackref=true,breaklinks=true,letterpaper=true,colorlinks,bookmarks=false]{hyperref}

\cvprfinalcopy 

\def\cvprPaperID{****} 

\ifcvprfinal\fi
\begin{document}

\title{RandAugment: Practical automated data augmentation\\ with a reduced search space}


\author{Ekin D. Cubuk \thanks{Authors contributed equally.}, Barret Zoph\footnotemark[1], Jonathon Shlens, Quoc V. Le  \\
Google Research, Brain Team\\
\texttt{\{cubuk, barretzoph, shlens, qvl\}@google.com}}

\maketitle

\input{abstract}

\input{intro}
\input{related}
\input{methods}

\input{results}
\input{conclusion}
\section{Acknowledgements}
We thank Samy Bengio, Daniel Ho, Ildoo Kim, Jaehoon Lee, Zhaoqi Leng, Hanxiao Liu, Raphael Gontijo Lopes, Ruoming Pang, Ben Poole, Mingxing Tan, and the rest of the Brain team for their help.
\clearpage


{\small

\input{paper.bbl}
\bibliographystyle{ieee_fullname}
}

\clearpage
\input{appendix}

\end{document}

%% file: abstract.tex
\begin{abstract}
Recent work has shown that data augmentation has the potential to significantly improve the generalization of deep learning models. Recently, automated augmentation strategies have led to state-of-the-art results in image classification and object detection. While these strategies were optimized for improving validation accuracy, they also led to state-of-the-art results in semi-supervised learning and improved robustness to common corruptions of images. 
An obstacle to a large-scale adoption of these methods is a separate search phase which increases the training complexity and may substantially increase the computational cost.
Additionally, due to the separate search phase, these approaches
are unable to adjust the regularization strength based on model or dataset size. Automated augmentation policies are often found by training small models on small datasets and subsequently applied to train larger models. In this work, we remove both of these obstacles. RandAugment has a significantly reduced search space which allows it to be trained on the target task with no need for a separate proxy task.  Furthermore, due to the parameterization, the regularization strength may be tailored to different model and dataset sizes.
RandAugment can be used uniformly across different tasks and datasets and works out of the box, matching or surpassing all previous automated augmentation approaches on CIFAR-10/100, SVHN, and ImageNet. On the ImageNet dataset we achieve 85.0\% accuracy, a 0.6\% increase over the previous state-of-the-art and 1.0\% increase over baseline augmentation. On object detection, RandAugment leads to 1.0-1.3\% improvement over baseline augmentation, and is within 0.3\% mAP of AutoAugment on COCO. Finally, due to its interpretable hyperparameter, RandAugment may be used to investigate the role of data augmentation with varying model and dataset size. Code is available online.
\footnote{\texttt{\href{}{github.com/tensorflow/tpu/tree/master/models/ \\ official/efficientnet}}}

\end{abstract}

%% file: intro.tex
\input{intro_table}

\section{Introduction}

Data augmentation is a widely used method for generating additional data to improve machine learning systems, for image classification~\cite{simard2003best,krizhevsky2012imagenet,devries2017dataset,zhang2017mixup}, object detection~\cite{girshick2018detectron}, instance segmentation~\cite{fang2019instaboost}, and speech recognition~\cite{kanda2013elastic, hannun2014deep, park2019specaugment}. Unfortunately, data augmentation methods require expertise, and manual work to design policies that capture prior knowledge in each domain. This requirement makes it difficult to extend existing data augmentation methods to other applications and domains.  

Learning policies for data augmentation 
has recently emerged as a method to automate the design of augmentation strategies and therefore has the potential to address some weaknesses of traditional data augmentation methods
\cite{cubuk2018autoaugment, zoph2019learning, ho2019population, lim2019fast}. Training a machine learning model with a learned data augmentation policy
may significantly improve accuracy~\cite{cubuk2018autoaugment}, model robustness~\cite{lopes2019improving, yin2019afourier, recht2018cifar}, and performance on semi-supervised learning~\cite{xie2019unsupervised} for image classification; likewise, for object detection tasks on COCO and PASCAL-VOC~\cite{zoph2019learning}. Notably, unlike engineering better network architectures  \cite{zoph2017learning}, all of these improvements in predictive performance incur no additional  computational cost at inference time.

In spite of the benefits of learned data augmentation policies, the computational requirements as well as the added complexity of two separate optimization procedures can be prohibitive.
The original presentation of neural architecture search (NAS) realized an analogous scenario in which the dual optimization procedure resulted in superior predictive performance, but the original implementation proved prohibitive in terms of complexity and computational demand.
Subsequent work accelerated training efficiency and the efficacy of the procedure \cite{liu2018darts,pham2018efficient,liu2017progressive,liu2017hierarchical}, eventually making the method amenable to a unified optimization based on a differentiable process \cite{liu2018darts}.
In the case of learned augmentations, subsequent work identified more efficient search methods \cite{ho2019population,lim2019fast}, however such methods still require a separate optimization procedure, which significantly increases the computational cost and complexity of training a machine learning model.

The original formulation for automated data augmentation postulated a separate search on a small, proxy task whose results may be transferred to a larger target task \cite{zoph2017learning, zoph2016neural}. This formulation makes a strong assumption that the proxy task provides a predictive indication of the larger task \cite{liu2017progressive, chen2018searching}. In the case of learned data augmentation, we provide experimental evidence to challenge this core assumption. In particular, we demonstrate that
this strategy is sub-optimal as the strength of the augmentation depends strongly on model and dataset size. These results suggest that an improved data augmentation may be possible if one could remove the separate search phase on a proxy task.

In this work, we propose a practical method for automated data augmentation -- termed {\em RandAugment } -- that does not require a separate search. In order to remove a separate search, we find it necessary to dramatically reduce the search space for data augmentation.
The reduction in parameter space is in fact so dramatic that simple grid search is sufficient to find a data augmentation policy that outperforms all learned augmentation methods that employ a separate search phase.
Our contributions can be summarized as follows:

\begin{itemize}
\item We demonstrate that the optimal strength of a data augmentation depends on the model size and training set size. This observation indicates that a separate optimization of an augmentation policy on a smaller proxy task may be sub-optimal for learning and transferring augmentation policies.

\item We introduce a vastly simplified search space for data augmentation containing 2 interpretable hyper-parameters. One may employ simple grid search to tailor the augmentation policy to a model and dataset, removing the need for a separate search process.

\item Leveraging this formulation, we demonstrate state-of-the-art results on CIFAR~\cite{krizhevsky2009learning}, SVHN~\cite{netzer2011reading}, and ImageNet~\cite{imagenet2009}. On object detection~\cite{lin2014microsoft}, our method is within 0.3\% mAP of state-of-the-art. On ImageNet we achieve a state-of-the-art accuracy of 85.0\%, a 0.6\% increment over previous methods and 1.0\% over baseline augmentation.
\end{itemize}

%% file: intro_table.tex
\bgroup
\def\arraystretch{1.3}
\begin{table}[t]
\scriptsize
\centering
\begin{tabular}{l|c|c|c|c|c}
&search &CIFAR-10 & SVHN & ImageNet& ImageNet\\
&space&PyramidNet & WRN & ResNet & E. Net-B7\\
\hline

Baseline & 0 & 97.3  & 98.5  & 76.3  & 84.0 \\
\hline
AA        & $10^{32}$              & 98.5                & 98.9                     & 77.6                          & 84.4            \\
Fast AA   & $10^{32}$               & 98.3                         & 98.8                    & 77.6                           & -                  \\
PBA       & $10^{61}$               & 98.5            & 98.9                  & -                   & -                  \\
RA (ours) & $10^{2\,\,\,}$                & 98.5                           & 99.0      & 77.6                             & 85.0             \\
\end{tabular}
\vspace{0.2cm}
\caption{\textbf{RandAugment matches or exceeds predictive performance of other augmentation methods with a significantly reduced search space.} We report the search space size and the test accuracy achieved for AutoAugment (AA) \cite{cubuk2018autoaugment}, Fast AutoAugment \cite{lim2019fast}, Population Based Augmentation (PBA) \cite{ho2019population} and the proposed RandAugment (RA) on CIFAR-10 \cite{krizhevsky2009learning}, SVHN \cite{netzer2011reading}, and ImageNet \cite{imagenet2009} classification tasks.
Architectures presented include PyramidNet \cite{han2017deep}, Wide-ResNet-28-10 \cite{WRN2016}, ResNet-50 \cite{he2016deep},  and EfficientNet-B7 \cite{tan2019efficientnet}.
Search space size is reported as the order of magnitude of the number of possible augmentation policies. All accuracies are the percentage on a cross-validated validation or test split. Dash indicates that results are not available.}
\label{tab:summary_results}  
\end{table}
\egroup

%% file: related.tex
\section{Related Work}

Data augmentation has played a central role in the training of deep vision models. On natural images, horizontal flips and random cropping or translations of the images are commonly used in classification and detection models~\cite{WRN2016,krizhevsky2012imagenet,girshick2018detectron}.  On MNIST, elastic distortions across scale, position, and orientation have been applied to achieve impressive results~\cite{simard2003best,ciregan2012multi,wan2013regularization,sato2015apac}. While previous examples augment the data while keeping it in the training set distribution, operations that do the opposite can also be effective in increasing generalization. Some methods randomly erase or add noise to patches of images for increased validation accuracy~\cite{cutout2017,zhong2017random}, robustness~\cite{szegedy2013intriguing,yin2019afourier,ford2019adversarial}, or both~\cite{lopes2019improving}.  Mixup~\cite{zhang2017mixup} is a particularly effective augmentation method on CIFAR-10 and ImageNet, where the neural network is trained on convex combinations of images and their corresponding labels. Object-centric cropping is commonly used for object detection tasks \cite{liu2016ssd}, whereas \cite{dwibedi2017cut} adds new objects on training images by cut-and-paste.

Moving away from individual operations to augment data, other work has focused on finding optimal strategies for combining different operations. For example, Smart Augmentation learns a network that merges two or more samples from the same class to generate new data~\cite{lemley2017smart}. Tran et al. generate augmented data via a Bayesian approach, based on the distribution learned from the training set~\cite{tran2017bayesian}. DeVries et al. use transformations (e.g. noise, interpolations and extrapolations) in the learned feature space to augment data~\cite{devries2017dataset}. Furthermore, generative adversarial networks (GAN) have been used to choose optimal sequences of data augmentation operations\cite{ratner2017learning}. GANs have also been used to generate training data directly~\cite{perez2017effectiveness,mun2017generative,zhu2017data,antoniou2017data,sixt2016rendergan}, however this approach does not seem to be as beneficial as learning sequences of data augmentation operations that are pre-defined~\cite{ravuri2019classification}.   

Another approach to learning data augmentation strategies from data is AutoAugment~\cite{cubuk2018autoaugment}, which originally used reinforcement learning to choose a sequence of operations as well as their probability of application and magnitude. Application of AutoAugment policies involves stochasticity at multiple levels: 1) for every image in every minibatch, a sub-policy is chosen with uniform probability. 2) operations in each sub-policy has an associated probability of application. 3) Some operations have stochasticity over direction. For example, an image can be rotated clockwise or counter-clockwise. The layers of stochasticity increase the amount of diversity that the network is trained on, which in turn was found to significantly improve generalization on many datasets. More recently, several papers used the AutoAugment search space and formalism with improved optimization algorithms to find AutoAugment policies more efficiently~\cite{ho2019population, lim2019fast}. Although the time it takes to search for policies has been reduced significantly, having to implement these methods in a separate search phase reduces the applicability of AutoAugment. For this reason, this work aims to eliminate the search phase on a separate proxy task completely.    

Some of the developments in RandAugment were inspired by the recent improvements to searching over data augmentation policies. For example, Population Based Augmentation (PBA) \cite{ho2019population} found that the optimal magnitude of augmentations increased during the course of training, which inspired us to not search over optimal magnitudes for each transformation but have a fixed magnitude schedule, which we discuss in detail in Section~\ref{sec:methods}. Furthermore, authors of Fast AutoAugment~\cite{lim2019fast} found that a data augmentation policy that is trained for density matching leads to improved generalization accuracy, which inspired our first order differentiable term for improving augmentation (see Section~\ref{first_order_term}). 

%% file: methods.tex
\section{Methods}
\label{sec:methods}

\begin{figure}[t]
\begin{center}
\centerline{\includegraphics[width=\linewidth]{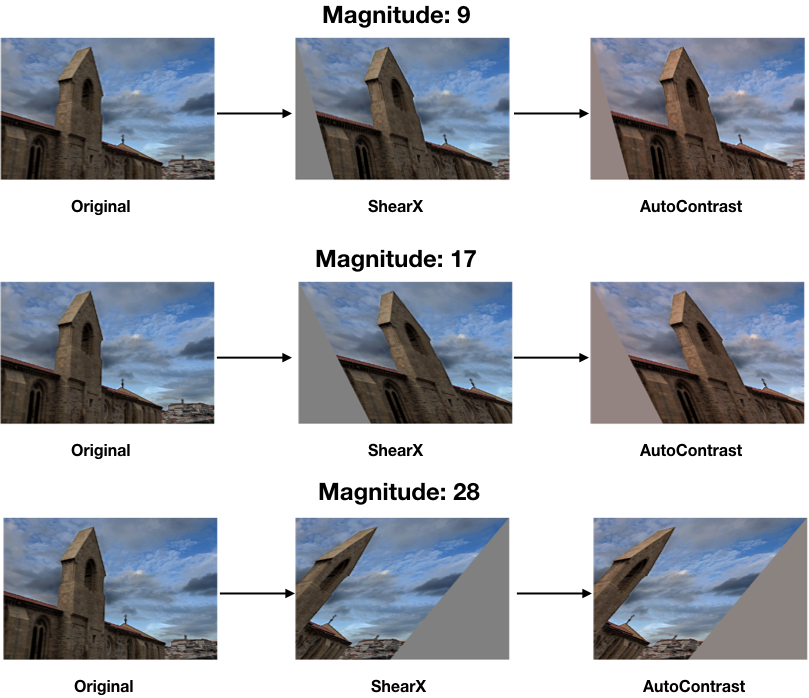}}
\caption{\textbf{Example images augmented by RandAugment.} In these examples $N$=2 and three magnitudes are shown corresponding to the optimal distortion magnitudes for ResNet-50, EfficientNet-B5 and EfficientNet-B7, respectively. As the distortion magnitude increases, the strength of the augmentation increases.}
\label{fig:imagenet_mag_images}
\end{center}
\vspace{-1.5cm}
\end{figure}

The primary goal of RandAugment is to remove the need for a separate search phase on a proxy task. The reason we wish to remove the search phase is because a separate search phase significantly complicates training and is computationally expensive.
More importantly, the proxy task may provide sub-optimal results (see Section~\ref{sec:failures_of_AA}). 
In order to remove a separate search phase, we aspire to fold the parameters for the data augmentation strategy into the hyper-parameters for training a model. Given that previous learned augmentation methods contained 30+ parameters \cite{cubuk2018autoaugment, lim2019fast, ho2019population}, we  focus on vastly reducing the parameter space for data augmentation.

Previous work indicates that the main benefit of learned augmentation policies arise from increasing the diversity of examples \cite{cubuk2018autoaugment,ho2019population,lim2019fast}. 
Indeed, previous work enumerated a policy in terms of choosing which transformations to apply out of $K$=14 available transformations, and probabilities for applying each transformation:
\begin{table}[H]
\footnotesize
\centering
\def\arraystretch{1.0}
\begin{tabular}{lll}
$\bullet\;$ \texttt{identity} & $\bullet\;$ \texttt{autoContrast} &
$\bullet\;$ \texttt{equalize} \\ $\bullet\;$ \texttt{rotate} &
$\bullet\;$ \texttt{solarize} & $\bullet\;$ \texttt{color} \\
$\bullet\;$ \texttt{posterize} &
$\bullet\;$ \texttt{contrast} & $\bullet\;$ \texttt{brightness} \\
$\bullet\;$ \texttt{sharpness} & $\bullet\;$ \texttt{shear-x} &
$\bullet\;$ \texttt{shear-y} \\
$\bullet\;$ \texttt{translate-x} &
$\bullet\;$ \texttt{translate-y} \\
\end{tabular}
\end{table}
\vspace{-0.4cm}
\noindent
In order to reduce the parameter space but still maintain image diversity, we replace the learned policies and probabilities for applying each transformation with a parameter-free procedure of {\em always} selecting a transformation with uniform probability $\frac{1}{K}$.
Given $N$ transformations for a training image, RandAugment may thus express $K^{N}$ potential policies.

\input{algorithm_figure}

The final set of parameters to consider is the magnitude of the each augmentation distortion.
Following \cite{cubuk2018autoaugment}, we employ the same linear scale for indicating the strength of each transformation. Briefly, each transformation resides on an integer scale from 0 to 10 where a value of 10 indicates the maximum scale for a given transformation. A data augmentation policy consists of identifying an integer for each augmentation \cite{cubuk2018autoaugment, lim2019fast, ho2019population}.
In order to reduce the parameter space further, we observe that the learned magnitude for each transformation follows a similar schedule during training (e.g. Figure 4 in \cite{ho2019population}) and postulate that a {\em single} global distortion $M$ may suffice for parameterizing all transformations. We experimented with four methods for the schedule of $M$ during training: constant magnitude, random magnitude, a linearly increasing magnitude, and a random magnitude with increasing upper bound. The details of this experiment can be found in Appendix~\ref{sec:magnitude_methods}. 

The resulting algorithm contains two parameters $N$ and $M$ and may be expressed simply in two lines of Python code (Figure \ref{fig:algorithm}).
Both parameters are human-interpretable such that larger values of $N$ and $M$ increase regularization strength.
Standard methods may be employed to efficiently perform hyperparameter optimization \cite{snoek2012practical, golovin2017google}, however given the extremely small search space we find that naive grid search is quite effective (Section \ref{sec:failures_of_AA}). We justify all of the choices of this proposed algorithm in this subsequent sections by comparing the efficacy of the learned augmentations to {\em all} previous learned data augmentation methods.

%% file: algorithm_figure.tex
\begin{figure}[t]
\begin{lstlisting}
transforms = [
'Identity', 'AutoContrast', 'Equalize', 
'Rotate', 'Solarize', 'Color', 'Posterize',
'Contrast', 'Brightness', 'Sharpness', 
'ShearX', 'ShearY', 'TranslateX', 'TranslateY']

def randaugment(N, M): 
"""Generate a set of distortions.

   Args:
     N: Number of augmentation transformations to apply sequentially.
     M: Magnitude for all the transformations.
"""

  sampled_ops = np.random.choice(transforms, N)
  return [(op, M) for op in sampled_ops]

\end{lstlisting}
\caption{Python code for RandAugment based on numpy.}
\label{fig:algorithm}
\end{figure}

%% file: results.tex
\section{Results}

\begin{figure*}[t]
\begin{center}
\centerline{\includegraphics[width=0.67\linewidth]{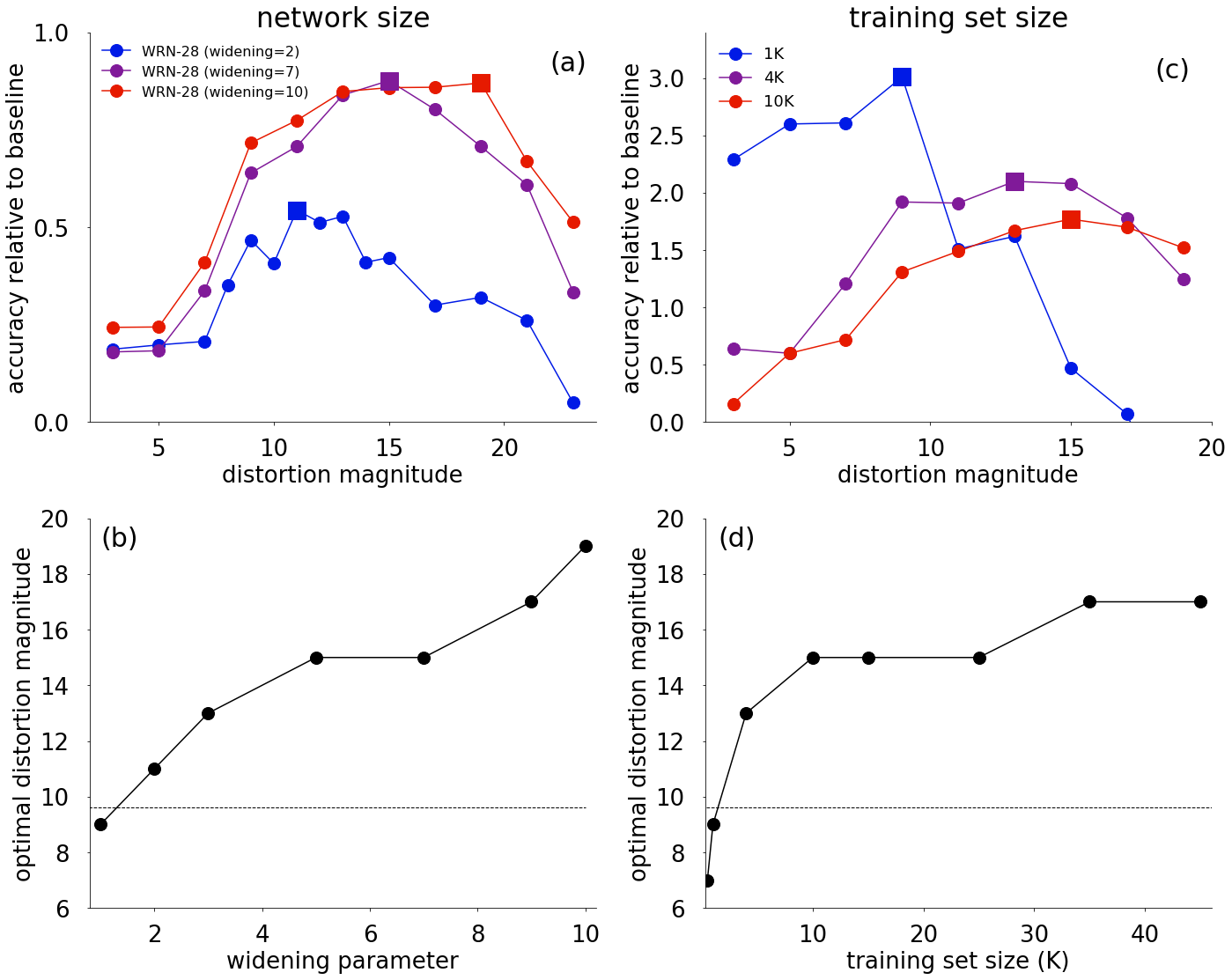}}
\caption{\textbf{Optimal magnitude of augmentation depends on the size of the model and the training set.} All results report CIFAR-10 validation accuracy for Wide-ResNet model architectures~\cite{WRN2016} averaged over 20 random initializations, where $N=1$.
(a) Accuracy of Wide-ResNet-28-2, Wide-ResNet-28-7, and Wide-ResNet-28-10 across varying distortion magnitudes. Models are trained for 200 epochs on 45K training set examples. 
Squares indicate the distortion magnitude that achieves the maximal accuracy.
(b) Optimal distortion magnitude across 7 Wide-ResNet-28 architectures with varying widening parameters ($k$). (c) Accuracy of Wide-ResNet-28-10 for three training set sizes (1K, 4K, and 10K) across varying distortion magnitudes. Squares indicate the distortion magnitude that achieves the maximal accuracy. (d) Optimal distortion magnitude across 8 training set sizes. Dashed curves show the scaled expectation value of the distortion magnitude in the AutoAugment policy \cite{cubuk2018autoaugment}.}
\label{fig:mag_vs}
\end{center}
\vspace{-1.0cm}
\end{figure*}

To explore the space of data augmentations, we experiment with core image classification and object detection tasks. In particular, we focus on CIFAR-10, CIFAR-100, SVHN, and ImageNet datasets as well as COCO object detection so that we may compare with previous work. For all of these datasets, we replicate the corresponding architectures and set of data transformations. Our goal is to demonstrate the relative benefits of employing this method over previous learned augmentation methods.

\input{small_table}

\subsection{Systematic failures of a separate proxy task}
\label{sec:failures_of_AA}
A central premise of learned data augmentation is to construct a small, proxy task that may be reflective of a larger task \cite{zoph2016neural, zoph2017learning, cubuk2018autoaugment}. Although this assumption is sufficient for identifying learned augmentation policies to improve performance \cite{cubuk2018autoaugment, zoph2019learning, park2019specaugment, lim2019fast, ho2019population}, it is unclear if this assumption is overly stringent and may lead to sub-optimal data augmentation policies.

In this first section, we challenge the hypothesis that formulating the problem in terms of a small proxy task is appropriate for learned data augmentation. In particular, we explore this question along two separate dimensions that are commonly restricted to achieve a small proxy task: model size and dataset size. To explore this hypothesis, we systematically measure the effects of data augmentation policies on CIFAR-10. First, we train a family of  Wide-ResNet architectures \cite{WRN2016}, where the model size may be systematically altered through the {\em widening} parameter governing the number of convolutional filters. For each of these networks, we train the model on CIFAR-10 and measure the final accuracy compared to a baseline model trained with default data augmentations (i.e. flip left-right and random translations). The Wide-ResNet models are trained with the additional $K$=14 data augmentations (see Methods) over a range of global distortion magnitudes $M$ parameterized on a uniform linear scale ranging from [0, 30] \footnote{Note that the range of magnitudes exceeds the specified range of magnitudes in the Methods because we wish to explore a larger range of magnitudes for this preliminary experiment. We retain the same scale as \cite{cubuk2018autoaugment} for a value of 10 to maintain comparable results.}.

Figure \ref{fig:mag_vs}a demonstrates the relative gain in accuracy of a model trained across increasing distortion magnitudes for three Wide-ResNet models. The squares indicate the distortion magnitude with which achieves the highest accuracy. Note that in spite of the measurement noise, Figure \ref{fig:mag_vs}a demonstrates systematic trends across distortion magnitudes. In particular, plotting all Wide-ResNet architectures versus the optimal distortion magnitude highlights a clear monotonic trend across increasing network sizes (Figure \ref{fig:mag_vs}b). Namely, larger networks demand larger data distortions for regularization. Figure \ref{fig:imagenet_mag_images} highlights the visual difference in the optimal distortion magnitude for differently sized models. Conversely, a learned policy based on \cite{cubuk2018autoaugment} provides a fixed distortion magnitude (Figure \ref{fig:mag_vs}b, dashed line) for all architectures that is clearly sub-optimal.

A second dimension for constructing a small proxy task is to train the proxy on a small subset of the training data.
Figure \ref{fig:mag_vs}c demonstrates the relative gain in accuracy of Wide-ResNet-28-10 trained across increasing distortion magnitudes for varying amounts of CIFAR-10 training data.
The squares indicate the distortion magnitude with that achieves the highest accuracy. Note that in spite of the measurement noise, Figure \ref{fig:mag_vs}c demonstrates systematic trends across distortion magnitudes.
We first observe that models trained on smaller training sets may gain more improvement from data augmentation  (e.g. 3.0\% versus 1.5\% in Figure \ref{fig:mag_vs}c). Furthermore, we see that the optimal distortion magnitude is larger for models that are trained on larger datasets. At first glance, this may disagree with the expectation that smaller datasets require stronger regularization.

Figure ~\ref{fig:mag_vs}d demonstrates that the optimal distortion magnitude increases monotonically with training set size.
One hypothesis for this counter-intuitive behavior is that aggressive data augmentation leads to a low signal-to-noise ratio in small datasets. Regardless, this trend highlights the need for increasing the strength of data augmentation on larger datasets and the shortcomings of optimizing learned augmentation policies on a proxy task comprised of a subset of the training data. Namely, the learned augmentation may learn an augmentation strength more tailored to the proxy task instead of the larger task of interest.

The dependence of augmentation strength on the dataset and model size indicate that a small proxy task may provide a sub-optimal indicator of performance on a larger task. This empirical result suggests that a distinct strategy may be necessary for finding an optimal data augmentation policy. In particular, we propose in this work to focus on a {\em unified} optimization of the model weights and data augmentation policy. Figure \ref{fig:mag_vs} suggest that merely searching for a shared distortion magnitude $M$ across all transformations may provide sufficient gains that exceed learned optimization methods \cite{cubuk2018autoaugment}. Additionally, we see that optimizing individual magnitudes further leads to minor improvement in performance (see Section~\ref{sec:individual_mag} in Appendix). 

Furthermore, Figure \ref{fig:mag_vs}a and \ref{fig:mag_vs}c indicate that merely sampling a few distortion magnitudes is sufficient to achieve good results. Coupled with a second free parameter $N$, we consider these results to prescribe an algorithm for learning an augmentation policy. In the subsequent sections, we identify two free parameters $N$ and $M$ specifying RandAugment through a minimal grid search and compare these results against computationally-heavy learned data augmentations based on proxy tasks.

\subsection{CIFAR}

CIFAR-10 has been extensively studied with previous data augmentation methods and we first test this proposed method on this data. The default augmentations for all methods include flips, pad-and-crop and Cutout~\cite{cutout2017}. $N$ and $M$ were selected based on the validation performance on 5K held out examples from the training set for 1 and 5 settings for $N$ and $M$, respectively. Results indicate that RandAugment achieves either competitive (i.e. within 0.1\%) or state-of-the-art on CIFAR-10 across four network architectures (Table~\ref{tab:small_results}). As a more challenging task, we additionally compare the efficacy of RandAugment on CIFAR-100 for Wide-ResNet-28-2 and Wide-ResNet-28-10. On the held out 5K dataset, we sampled 2 and 4 settings for $N$ and $M$, respectively (i.e. 
$N$=$\{1,2\}$ and $M$=$\{2, 6, 10, 14\}$). For Wide-ResNet-28-2 and Wide-ResNet-28-10, we find that $N$=1, $M$=2 and $N$=2, $M$=14 achieves best results, respectively.
Again, RandAugment achieves competitive or superior results across both architectures (Table~\ref{tab:small_results}).

\subsection{SVHN}

Because SVHN is composed of numbers instead of natural images, the data augmentation strategy for SVHN may differ substantially from CIFAR-10.
Indeed, \cite{cubuk2018autoaugment} identified a qualitatively different policy for CIFAR-10 than SVHN. Likewise, in a semi-supervised setting for CIFAR-10, a policy learned from CIFAR-10 performs better than a policy learned from SVHN ~\cite{xie2019unsupervised}.

SVHN has a core training set of 73K images \cite{netzer2011reading}. In addition, SVHN contains 531K less difficult ``extra'' images to augment training.
We compare the performance of the augmentation methods on SVHN with and without the extra data on Wide-ResNet-28-2 and Wide-ResNet-28-10 (Table \ref{tab:small_results}). In spite of the large differences between SVHN and CIFAR, RandAugment consistently matches or outperforms previous methods with no alteration to the list of transformations employed. Notably, for Wide-ResNet-28-2, applying RandAugment to the core training dataset improves performance more than augmenting with 531K additional training images (98.3\% vs. 98.2\%). For, Wide-ResNet-28-10, RandAugment is competitive with augmenting the core training set with 531K training images (i.e. within 0.2\%). Nonetheless, Wide-ResNet-28-10 with RandAugment matches the previous state-of-the-art accuracy on SVHN which used a more advanced model \cite{cubuk2018autoaugment}.

\subsection{ImageNet}
\input{imagenet_table}

Data augmentation methods that improve CIFAR-10 and SVHN models do not always improve large-scale tasks such as ImageNet. For instance, Cutout substantially improves CIFAR and SVHN performance~\cite{cutout2017}, but fails to improve ImageNet \cite{lopes2019improving}. Likewise, AutoAugment does not increase the performance on ImageNet as much as other tasks \cite{cubuk2018autoaugment}, especially for large networks (e.g. +0.4\% for AmoebaNet-C \cite{cubuk2018autoaugment} and +0.1\% for EfficientNet-B5 \cite{tan2019efficientnet}).
One plausible reason for the lack of strong gains is that the small proxy task was particularly impoverished by restricting the task to $\sim$10\% of the 1000 ImageNet classes.

Table~\ref{tab:imagenet_results} compares the performance of RandAugment to other learned augmentation approaches on ImageNet. RandAugment matches the performance of AutoAugment and Fast AutoAugment on the smallest model (ResNet-50), but on larger models RandAugment significantly outperforms other methods achieving increases of up to +1.3\% above the baseline. For instance, on EfficientNet-B7, the resulting model achieves 85.0\% -- a new state-of-the-art accuracy -- exhibiting a 1.0\% improvement over the baseline augmentation. These systematic gains are similar to the improvements achieved with engineering new architectures \cite{zoph2017learning,liu2017progressive}, however these gains arise without incurring  additional computational cost at inference time. 

\subsection{COCO}
To further test the generality of this approach, we next explore a related task of large-scale object detection on the COCO dataset \cite{lin2014microsoft}. 
Learned augmentation policies have improved object detection and lead to state-of-the-art results \cite{zoph2019learning}. We followed previous work by training on the same architectures and following the same training schedules (see Appendix~\ref{coco_details}). Briefly, we employed RetinaNet \cite{lin2017focal} with ResNet-101 and ResNet-200 as a backbone \cite{he2016deep}. Models were trained for 300 epochs from random initialization.

Table~\ref{tab:coco_results} compares results between a baseline model, AutoAugment and RandAugment. AutoAugment leveraged additional, specialized transformations not afforded to RandAugment in order to augment the localized bounding box of an image \cite{zoph2019learning}. In addition, note that AutoAugment expended $\sim$15K GPU hours for search, where as RandAugment was tuned by on merely 6 values of the hyper-parameters (see Appendix \ref{coco_details}). In spite of the smaller library of specialized transformations and the lack of a separate search phase, RandAugment surpasses the baseline model and provides competitive accuracy with AutoAugment. We reserve for future work to expand the transformation library to include bounding box specific transformation to potentially improve RandAugment results even further.

\begin{table}[t]
\centering
\small
\begin{tabular}{llcc}
  \hline
  model & augmentation &  mAP  & search space\\  
  \hline 
  &Baseline  & 38.8 & 0\\
  ResNet-101 &AutoAugment  & \textbf{40.4}& $10^{34}$\\
  &RandAugment & 40.1 & $10^{2\,\,\,}$\\
  \hline
   &Baseline  & 39.9 & 0\\
  ResNet-200 &AutoAugment  & \textbf{42.1} & $10^{34}$\\
  &RandAugment & 41.9 & $10^{2\,\,\,}$\\
  \hline 
\end{tabular}
\caption{\textbf{Results on object detection.} Mean average precision (mAP) on COCO detection task. Higher is better. Search space size is reported as the order of magnitude of the number of possible augmentation policies. Models are trained for 300 epochs from random initialization following \cite{zoph2019learning}.}
\label{tab:coco_results}  
\end{table}

\subsection{Investigating the dependence on the included transformations}
\begin{figure}[t]
\begin{center}
\centerline{\includegraphics[width=\linewidth]{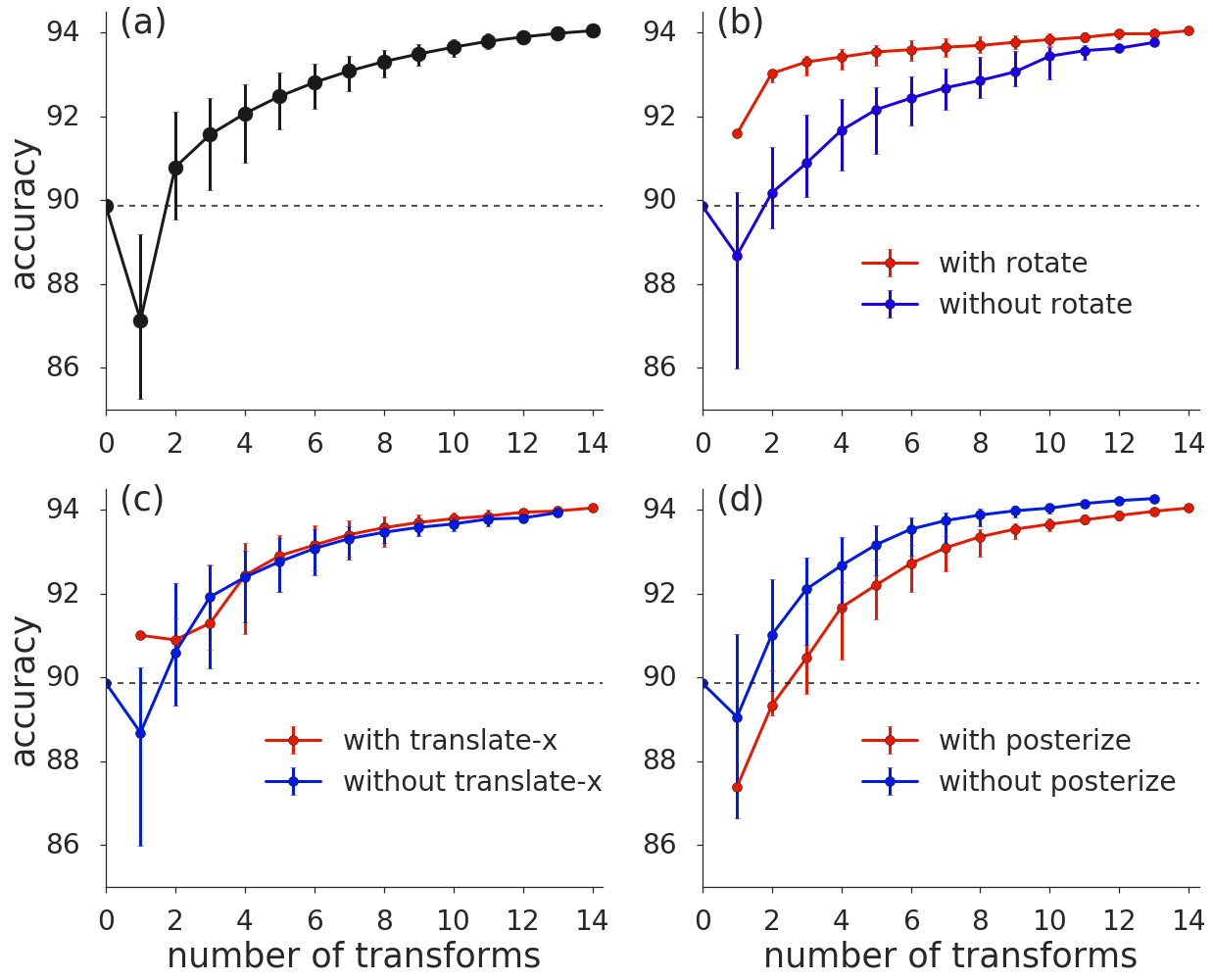}}
\caption{\textbf{Average performance improves when more transformations are included in RandAugment.} All panels report median CIFAR-10 validation accuracy for Wide-ResNet-28-2 model architectures~\cite{WRN2016} trained with RandAugment ($N=3$, $M=4$) using randomly sampled subsets of transformations. No other data augmentation is included in training. Error bars indicate 30\textsuperscript{th} and 70\textsuperscript{th} percentile. (a) Median accuracy for randomly sampled subsets of transformations. (b) Median accuracy for subsets with and without the {\footnotesize\texttt{Rotate}} transformation. (c) Median accuracy for subsets with and without the {\footnotesize\texttt{translate-x}} transformation. (d) Median accuracy for subsets with and without the {\footnotesize\texttt{posterize}} transformation. Dashed curves show the accuracy of the model trained without any augmentations.}
\label{fig:ablation_sampling}
\end{center}
\vspace{-0.7cm}
\end{figure}

RandAugment achieves state-of-the-art results across different tasks and datasets using the same list of transformations. This result suggests that RandAugment is largely insensitive to the selection of transformations for different datasets. To further study the sensitivity, we experimented with RandAugment on a Wide-ResNet-28-2 trained on CIFAR-10 for randomly sampled subsets of the full list of 14 transformations. We did not use flips, pad-and-crop, or cutout to only focus on the improvements due to RandAugment with random subsets. Figure~\ref{fig:ablation_sampling}a suggests that the median validation accuracy due to RandAugment improves as the number of transformations is increased. However, even with only two transformations, RandAugment leads to more than 1\% improvement in validation accuracy on average. 

To get a sense for the effect of individual transformations, we calculate the average improvement in validation accuracy for each transformation when they are added to a random subset of transformations. We list the transformations in order of most helpful to least helpful in Table~\ref{tab:ablation_transform_list}. We see that while geometric transformations individually make the most difference, some of the color transformations lead to a degradation of validation accuracy on average. Note that while Table~\ref{tab:ablation_transform_list} shows the average effect of adding individual transformations to randomly sampled subsets of transformations, Figure~\ref{fig:ablation_sampling}a shows that including all transformations together leads to a good result. The transformation {\footnotesize\texttt{rotate}} is most helpful on average, which was also observed previously~\cite{cubuk2018autoaugment, zoph2019learning}. To see the effect of representative transformations in more detail, we repeat the analysis in Figure~\ref{fig:ablation_sampling}a for subsets with and without ({\footnotesize\texttt{rotate, translate-x, and posterize}}). Surprisingly, {\footnotesize\texttt{rotate}} can significantly improve performance and lower variation even when included in small subsets of RandAugment transformations, while {\footnotesize\texttt{posterize}} seems to hurt all subsets of all sizes.    

\begin{table}[t]
\centering
\small
\begin{tabular}{lr|lr}
 \hline
transformation & $\Delta$ (\%) &transformation & $\Delta$ (\%)\\
\hline
rotate&\textbf{1.3}&shear-x&0.9\\
shear-y&0.9&translate-y&0.4\\
translate-x&0.4&autoContrast&0.1\\
sharpness&0.1&identity&0.1\\
contrast&0.0&color&0.0\\
brightness&0.0&equalize&-0.0\\
solarize&-0.1&posterize&-0.3\\
\end{tabular}
\vspace{0.2cm}
\caption{{\bf Average improvement due to each transformation.} Average difference in validation accuracy (\%) when a particular transformation is added to a randomly sampled set of transformations. For this ablation study, Wide-ResNet-28-2 models were trained on CIFAR-10 using RandAugment ($N=3$, $M=4$) with the randomly sampled set of transformations, with no other data augmentation.}
\label{tab:ablation_transform_list}
\end{table}

\subsection{Learning the probabilities for selecting image transformations}
\label{first_order_term}

\begin{table}[t]
\centering
\small
\begin{tabular}{l|cc|cc}
    & \footnotesize{baseline} &   AA    &  RA & + 1\textsuperscript{st} \\
  \hline  
  \textbf{Reduced CIFAR-10} &&&&\\
  Wide-ResNet-28-2 &  82.0 & \textbf{85.6} & 85.3 & 85.5\\ 
  Wide-ResNet-28-10 & 83.5&  \textbf{87.7} & 86.8 & 87.4\\ 
  \hline 
  \textbf{CIFAR-10} &&&&\\
  Wide-ResNet-28-2 & 94.9  & 95.9 & 95.8 & \textbf{96.1}\\ 
  Wide-ResNet-28-10 & 96.1&  \textbf{97.4} & 97.3 & \textbf{97.4}\\ 
  \hline
\end{tabular}
\vspace{0.2cm}
\caption{\textbf{Differentiable optimization for augmentation can improve RandAugment.} Test accuracy (\%) from differentiable RandAugment for reduced (4K examples) and full CIFAR-10. The 1\textsuperscript{st}-order approximation (1\textsuperscript{st}) is based on density matching (Section \ref{first_order_term}). Models trained on reduced CIFAR-10 were trained for 500 epochs. CIFAR-10 models trained using the same hyperparameters as previous. Each result is averaged over 10 independent runs.}
\label{tab:diff_results}  
\vspace{-0.5cm}
\end{table}

RandAugment selects all image transformations with equal probability. This opens up the question of whether learning $K$ probabilities may improve performance further. Most of the image transformations (except 
{\footnotesize {\texttt posterize, equalize, and autoContrast}}) 
are differentiable, which permits back-propagation to learn the $K$ probabilities \cite{liu2018darts}.
Let us denote $\alpha_{ij}$ as the learned probability of selecting image transformation $i$ for operation $j$.
For $K$=14 image transformations and $N$=2 operations, $\alpha_{ij}$ constitutes 28 parameters. We initialize all weights such that each transformation is equal probability (i.e. RandAugment), and update these parameters based on how well a model classifies a held out set of validation images distorted by $\alpha_{ij}$.
This approach was inspired by density matching \cite{lim2019fast}, but instead uses a differentiable approach in  lieu of Bayesian optimization. We label this method as a 1\textsuperscript{st}-order density matching approximation.

To test the efficacy of density matching to learn the probabilities of each transformation, we trained Wide-ResNet-28-2 and Wide-ResNet-28-10 on CIFAR-10 and the reduced form of CIFAR-10 containing 4K training samples. Table~\ref{tab:diff_results} indicates that learning the probabilities $\alpha_{ij}$ slightly improves performance on
reduced and full CIFAR-10 (RA vs 1\textsuperscript{st}). The 1\textsuperscript{st}-order method improves accuracy by more than 3.0\% for both models on reduced CIFAR-10 compared to the baseline of flips and pad-and-crop. On CIFAR-10, the 1\textsuperscript{st}-order method improves accuracy by 0.9\% on the smaller model and 1.2\% on the larger model compared to the baseline. We further see that the 1\textsuperscript{st}-order method always performs better than RandAugment, with the largest improvement on Wide-ResNet-28-10 trained on reduced CIFAR-10 (87.4\% vs. 86.8\%). On CIFAR-10, the 1\textsuperscript{st}-order method outperforms AutoAugment on Wide-ResNet-28-2 (96.1\% vs. 95.9\%) and matches AutoAugment on Wide-ResNet-28-10 \footnote{As a baseline comparison, in preliminary experiments we additionally learn $\alpha_{ij}$  based on differentiating through a virtual training step \cite{liu2018darts}. In this approach, the 2\textsuperscript{nd}-order approximation yielded consistently negative results (see Appendix \ref{second_order_explained}).}.
Although the density matching approach is promising, this method can be expensive as one must apply all $K$ transformations $N$ times to each image independently. Hence, because the computational demand of $KN$ transformations is prohibitive for large images, we reserve this for future exploration.
In summary, we take these results to indicate that learning the probabilities through density matching may improve the performance on small-scale tasks and reserve explorations to larger-scale tasks for the future.

%% file: small_table.tex
\begin{table}[t]
\centering
\footnotesize
\begin{tabular}{l|cccc|c}
  & {\small baseline}        & PBA & Fast AA &   AA    &  RA \\
  \hline 
  \textbf{CIFAR-10} &&&&&\\
Wide-ResNet-28-2 & 94.9 & - & - & \textbf{95.9} & 95.8  \\ 
  Wide-ResNet-28-10 & 96.1 & \textbf{97.4} & 97.3 & \textbf{97.4} & 97.3  \\ 
  Shake-Shake  & 97.1 & \textbf{98.0} & \textbf{98.0} & \textbf{98.0} & \textbf{98.0}  \\ 
  PyramidNet  & 97.3 & \textbf{98.5} & 98.3 & \textbf{98.5} & \textbf{98.5} \\ 
  \hline
  \textbf{CIFAR-100} &&&&&\\
  Wide-ResNet-28-2 & 75.4 & - & - & \textbf{78.5} & 78.3 \\  
  Wide-ResNet-28-10 & 81.2 & \textbf{83.3} & 82.7 & 82.9 & \textbf{83.3}  \\ 
  \hline  
  \textbf{SVHN (core set)} &&&&&\\
  Wide-ResNet-28-2 & 96.7 & -  & - & 98.0 & \textbf{98.3}  \\ 
  Wide-ResNet-28-10 & 96.9 & -  & - & 98.1 & \textbf{98.3}  \\ 
  \hline  
  \textbf{SVHN} &&&&&\\
  Wide-ResNet-28-2 & 98.2 & - & - & \textbf{98.7}  & \textbf{98.7} \\
  Wide-ResNet-28-10 & 98.5 &98.9  & 98.8 & 98.9 & \textbf{99.0}  \\ 
  \hline
\end{tabular}
\vspace{0.2cm}
\caption{\textbf{Test accuracy (\%) on CIFAR-10, CIFAR-100, SVHN and SVHN core set}. Comparisons across default data augmentation (baseline), Population Based Augmentation (PBA)~\cite{ho2019population} and Fast AutoAugment (Fast AA)~\cite{lim2019fast}, AutoAugment (AA) \cite{cubuk2018autoaugment} and proposed RandAugment (RA). Note that baseline and AA are replicated in this work. SVHN core set consists of ~73K examples. The Shake-Shake model ~\cite{gastaldi2017shake} employed a 26 2$\times$96d configuration, and the
PyramidNet model used the ShakeDrop regularization~\cite{yamada2018shakedrop}. Results reported by us are averaged over 10 independent runs. Bold indicates best results.}
\label{tab:small_results}  
\end{table}

%% file: imagenet_table.tex
\begin{table*}[!ht]
\centering
\small
\begin{tabular}{l|cccc}
   & baseline        & Fast AA &   AA    &  RA \\
  \hline 
  ResNet-50 & 76.3 / 93.1 & \textbf{77.6} / 93.7 & \textbf{77.6} / \textbf{93.8} & \textbf{77.6} / \textbf{93.8}  \\ 
  EfficientNet-B5 & 83.2 / 96.7 & - & 83.3 / 96.7& \textbf{83.9 / 96.8}\\ 
  EfficientNet-B7 & 84.0 / 96.9 & - & 84.4 / 97.1 & \textbf{85.0 / 97.2}\\ 
  \hline
\end{tabular}
\vspace{0.2cm}
\caption{\textbf{ImageNet results.} Top-1 and Top-5 accuracies (\%) on ImageNet. Baseline and AutoAugment (AA) results on ResNet-50 are from ~\cite{cubuk2018autoaugment}.
Fast AutoAugment (Fast AA) results are from ~\cite{lim2019fast}. EfficientNet results with and without AutoAugment are from \cite{tan2019efficientnet}. Highest accuracy for each model is presented in bold. Note that Population Based Augmentation (PBA)~\cite{ho2019population} has not been implemented on ImageNet.}
\label{tab:imagenet_results}  
\end{table*}

%% file: conclusion.tex
\section{Discussion}
Data augmentation is a necessary method for achieving state-of-the-art performance \cite{simard2003best,krizhevsky2012imagenet,devries2017dataset,zhang2017mixup,girshick2018detectron,park2019specaugment}. Learned data augmentation strategies have helped automate the design of such strategies and likewise achieved state-of-the-art results \cite{cubuk2018autoaugment,lim2019fast,ho2019population,zoph2019learning}.
In this work, we demonstrated that previous methods of learned augmentation suffers from systematic drawbacks. Namely, not tailoring the number of distortions and the distortion magnitude to the dataset size nor the model size leads to sub-optimal performance.
To remedy this situation, we propose a  simple parameterization for targeting augmentation to particular model and dataset sizes. We demonstrate that RandAugment is competitive with or outperforms previous approaches \cite{cubuk2018autoaugment,lim2019fast,ho2019population,zoph2019learning} on CIFAR-10/100, SVHN, ImageNet and COCO without a separate search for data augmentation policies.

In previous work, scaling learned data augmentation to larger dataset and models have been a notable obstacle. For example, AutoAugment and Fast AutoAugment could only be optimized for small models on reduced subsets of data \cite{cubuk2018autoaugment,lim2019fast}; population based augmentation was not reported for large-scale problems \cite{ho2019population}. The proposed method scales quite well to datasets such as ImageNet and COCO while incurring minimal computational cost (e.g. 2 hyper-parameters), but notable predictive performance gains. An open question remains how this method may improve model robustness \cite{lopes2019improving, yin2019afourier, recht2018cifar} or semi-supervised learning  \cite{xie2019unsupervised}. Future work will study how this method applies to other machine learning domains, where data augmentation is known to improve predictive performance, such as image segmentation \cite{chen2017deeplab}, 3-D perception \cite{ngiam2019starnet}, speech recognition \cite{hinton2012deep} or audio recognition \cite{hershey2017cnn}. In particular, we wish to better understand if or when datasets or tasks may require a separate search phase to achieve optimal performance. Finally, an open question remains how one may tailor the set of transformations to a given tasks in order to further improve the predictive performance of a given model.

%% file: appendix.tex
\appendix

\section{Appendix}
\subsection{Second order term from bilevel optimization}
\label{second_order_explained}
For the second order term for the optimization of augmentation parameters, we follow the formulation in  \cite{liu2018darts}, which we summarize below. We treat the optimization of augmentation parameters and weights of the neural network as a bilevel optimization problem, where $\alpha$ are the augmentation parameters and $w$ are the weights of the neural network. Then the goal is to find the optimal augmentation parameters $\alpha$ such that when weights are optimized on the training set using data augmentation given by $\alpha$ parameters, the validation loss is minimized. In other words:
\begin{eqnarray}
    \lefteqn{\mathrm{min}_{\alpha} \mathcal{L}_{val}(w^*(\alpha),\alpha) \; \mathrm{s.t.} \; w^*(\alpha) =} \nonumber \\ &&\mathrm{argmin}_w \;\mathcal{L}_{train}(w,\alpha).
\end{eqnarray}
Then, again following \cite{liu2018darts}, we approximate this bilevel optimization by a single virtual training step,
\begin{eqnarray}
\lefteqn{\nabla_{\alpha} \mathcal{L}_{val} ( w^*(\alpha),\alpha) \approx} \nonumber \\ &&\nabla_{\alpha} \mathcal{L}_{val} ( w - \xi \nabla_w \mathcal{L}_{train} (w,\alpha), \alpha),
\label{virtual_step}
\end{eqnarray}
where $\xi$ is the virtual learning rate. Eq.~\ref{virtual_step} can be expanded as
\begin{eqnarray}
\lefteqn{\nabla_{\alpha} \mathcal{L}_{val} ( w^*(\alpha),\alpha) \approx}\nonumber \\&&\nabla_{\alpha} \mathcal{L}_{val} ( w - \xi \nabla_w \mathcal{L}_{train} (w,\alpha), \alpha)- \nonumber \\ &&\xi \nabla^2_{\alpha,w} \mathcal{L}_{train} ( w, \alpha) \nabla_{w'}\mathcal{L}_{val}(w',\alpha),
\label{chain_rule}
\end{eqnarray}

where $w' = w - \xi \nabla_w \mathcal{L}_{train} (w,\alpha)$. In the case where the virtual learning rate, $\xi$, is zero, the second term disappears and the first term becomes $\nabla \mathcal{L}_{val}(w,\alpha)$, which was called the first-order approximation \cite{liu2018darts}. This first-order approximation was found to be highly significant for architecture search, where most of the improvement (0.3\% out of 0.5\%) could be achieved using this approximation in a more efficient manner (1.5 days as opposed to 4 days). Unfortunately, when $\alpha$ represents augmentation parameters, first-order approximation is irrelevant since the predictions of a model on the clean validation images do not depend on the augmentation parameters $\alpha$. Then we are left with just the second order approximation, where $\xi>0$, which we approximate via finite difference approximation as
\begin{eqnarray}
\lefteqn{\nabla^2_{\alpha,w} \mathcal{L}_{train} ( w, \alpha) \nabla_{w'}\mathcal{L}_{val}(w',\alpha) \approx}\nonumber \\&&\frac{\nabla_\alpha \mathcal{L}_{train}(w^+, \alpha) - \nabla_\alpha \mathcal{L}_{train}(w^-, \alpha) }{2\epsilon},
\label{finite_difference}
\end{eqnarray}
where $w^{\pm} = w \pm \epsilon \nabla_{w'}\mathcal{L}_{val}(w', \alpha)$ and $\epsilon$ is a small number. 

\subsubsection{Magnitude methods}
\label{sec:magnitude_methods}
\begin{table}[t]
\centering
\small
\begin{tabular}{ll}
  \hline
  Magnitude Method & Accuracy \\
  
  \hline 
  Random Magnitude & 97.3 \\
  Constant Magnitude & 97.2 \\
  Linearly Increasing Magnitude & 97.2 \\
  Random Magnitude with Increasing Upper Bound & 97.3 \\
  \hline 
\end{tabular}
\vspace{0.2cm}
\caption{\textbf{Results for different ways of setting the global magnitude parameter $M$.} All magnitude methods were run on CIFAR-10 with Wide-ResNet-28-10 for 200 epochs. The reported accuracy is the average of 10 runs on the validation set for the best hyperparamter setting for that magnitude method. All magnitude methods searched over had 48 different hyperparameter settings tried.}
\label{tab:magnitude_results}  
\end{table}

A random magnitude uniformly randomly samples the distortion magnitude between two values. A constant magnitude sets the distortion magnitude to a constant number during the course of training. A linearly increasing magnitude interpolates the distortion magnitude during training between two values. A random magnitude with increasing upper bound is similar to a random magnitude, but the upper bound is increased linearly during training.
In preliminary experiments, we found that all strategies worked equally well. Thus, we selected a constant magnitude because this strategy includes only a single hyper-parameter, and we employ this for the rest of the work. The results from our experiment on trying the different magnitude strategies can be see in Table~\ref{tab:magnitude_results}.

\subsubsection{Optimizing individual transformation magnitudes}
\label{sec:individual_mag}
\begin{figure}[t]
\begin{center}
\centerline{\includegraphics[width=\linewidth]{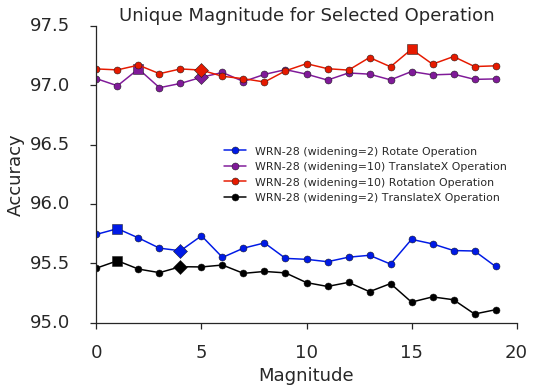}}
\caption{\textbf{Performance when magnitude is changed for one image transformation.} This plot uses a shared magnitude for all image transformations and then changes the magnitude of only one operation while keeping the others fixed. Two different architectures were tried (WRN-28-2 and WRN-28-10) and two different image transformations were changed (Rotate and TranslateX), which results in the 4 lines shown. Twenty different magnitudes were tried for the selected transformation ($[0-19]$). The squares indicate the optimal magnitude found and the diamonds indicate the magnitude used for all other transformations (4 for WRN-28-2 and 5 for WRN-28-10). }
\label{fig:mag_ablation}
\end{center}
\vspace{-1.0cm}
\end{figure}

Figure~\ref{fig:mag_ablation} demonstrates that changing the magnitude for one transformation, when keeping the rest fixed results in a very minor accuracy change. This suggests that tying all magnitudes together into a single value $M$ is not greatly hurting the model performance. Across all for settings in Figure~\ref{fig:mag_ablation} the difference in accuracy of the tied magnitude vs the optimal one found was 0.19\% 0.18\% for the rotation operation experiments and 0.07\% 0.05\% for the TranslateX experiments. Changing one transformation does not have a huge impact on performance, which leads us to think that tying all magnitude parameters together is a sensible approach that drastically reduces the size of the search-space.

\subsection{Experimental Details}
\label{experimental_details}
\subsubsection{CIFAR}
The Wide-ResNet models were trained for 200 epochs with a learning rate of 0.1, batch size of 128, weight decay of 5e-4, and cosine learning rate decay. Shake-Shake~\cite{gastaldi2017shake} model was trained for 1800 epochs with a learning rate of 0.01, batch size of 128, weight decay of 1e-3, and cosine learning rate decay. ShakeDrop~\cite{yamada2018shakedrop} models were trained for 1800 epochs with a learning rate of 0.05, batch size of 64 (as 128 did not fit on a single GPU), weight decay of 5e-5, and cosine learning rate decay.   

On CIFAR-10, we used 3 for the number of operations applied ($N$) and tried 4, 5, 7, 9, and 11 for magnitude. For Wide-ResNet-2 and Wide-ResNet-10, we find that the optimal magnitude is 4 and 5, respectively. For Shake-Shake (26 2x96d) and PyramidNet + ShakeDrop models, the optimal magnitude was 9 and 7, respectively.   
\subsubsection{SVHN}
For both SVHN datasets, we applied cutout after RandAugment as was done for AutoAugment and related methods. On core SVHN, for both Wide-ResNet-28-2 and Wide-ResNet-28-10, we used a learning rate of 5e-3, weight decay of 5e-3, and cosine learning rate decay for 200 epochs. We set $N=3$ and tried 5, 7, 9, and 11 for magnitude. For both Wide-ResNet-28-2 and Wide-ResNet-28-10, we find the optimal magnitude to be 9.  

On full SVHN, for both Wide-ResNet-28-2 and Wide-ResNet-28-10, we used a learning rate of 5e-3, weight decay of 1e-3, and cosine learning rate decay for 160 epochs. We set $N=3$ and tried 5, 7, 9, and 11 for magnitude. For Wide-ResNet-28-2, we find the optimal magnitude to be 5; whereas for Wide-ResNet-28-10, we find the optimal magnitude to be 7.  
\subsubsection{ImageNet}
The ResNet models were trained for 180 epochs using the standard ResNet-50 training hyperparameters. The image size was 224 by 244, the weight decay was 0.0001 and the momentum optimizer with a momentum parameter of 0.9 was used. The learning rate was 0.1, which gets scaled by the batch size divided by 256.
A global batch size of 4096 was used, split across 32 workers. For ResNet-50 the optimal distortion magnitude was 9 and ($N=2$). The distortion magnitudes we tried were 5, 7, 9, 11, 13, 15 and the values of $N$ that were tried were 1, 2 and 3.

The EfficientNet experiments used the default hyper parameters and training schedule, which can be found in ~\cite{tan2019efficientnet}. We trained for 350 epochs, used a batch size of 4096 split across 256 replicas. The learning rate was 0.016, which gets scaled by the batch size divided by 256. We used the RMSProp optimizer with a momentum rate of 0.9, epsilon of 0.001 and a decay of 0.9. The weight decay used was 1e-5. For EfficientNet B5 the image size was 456 by 456 and for EfficientNet B7 it was 600 by 600. For EfficientNet B5 we tried $N=2$ and $N=3$ and found them to perform about the same. We found the optimal distortion magnitude for B5 to be 17. The different magnitudes we tried were 8, 11, 14, 17, 21. For EfficientNet B7 we used $N=2$ and found the optimal distortion magnitude to be 28. The magnitudes tried were 17, 25, 28, 31.

The default augmentation of horizontal flipping and random crops were used on ImageNet, applied before RandAugment. The standard training and validation splits were employed for training and evaluation.

\subsection{COCO}
\label{coco_details}
We applied horizontal flipping and scale jitters in addition to RandAugment. We used the same list of data augmentation transformations as we did in all other classification tasks. Geometric operations transformed the bounding boxes the way it was defined in Ref.~\cite{zoph2019learning}. We used a learning rate of 0.08 and a weight decay of 1e − 4. The focal loss parameters are set to be $\alpha = 0.25$ and $\gamma = 1.5$. We set $N=1$ and tried distortion magnitudes between 4 and 9. We found the optimal distortion magnitude for ResNet-101 and ResNet-200 to be 5 and 6, respectively. 

%% file: paper.bbl
\begin{thebibliography}{10}\itemsep=-1pt

\bibitem{antoniou2017data}
Antreas Antoniou, Amos Storkey, and Harrison Edwards.
\newblock Data augmentation generative adversarial networks.
\newblock {\em arXiv preprint arXiv:1711.04340}, 2017.

\bibitem{chen2018searching}
Liang-Chieh Chen, Maxwell Collins, Yukun Zhu, George Papandreou, Barret Zoph,
  Florian Schroff, Hartwig Adam, and Jon Shlens.
\newblock Searching for efficient multi-scale architectures for dense image
  prediction.
\newblock In {\em Advances in Neural Information Processing Systems}, pages
  8699--8710, 2018.

\bibitem{chen2017deeplab}
Liang-Chieh Chen, George Papandreou, Iasonas Kokkinos, Kevin Murphy, and Alan~L
  Yuille.
\newblock Deeplab: Semantic image segmentation with deep convolutional nets,
  atrous convolution, and fully connected crfs.
\newblock {\em IEEE transactions on pattern analysis and machine intelligence},
  40(4):834--848, 2017.

\bibitem{ciregan2012multi}
Dan Ciregan, Ueli Meier, and J{\"u}rgen Schmidhuber.
\newblock Multi-column deep neural networks for image classification.
\newblock In {\em Proceedings of IEEE Conference on Computer Vision and Pattern
  Recognition}, pages 3642--3649. IEEE, 2012.

\bibitem{cubuk2018autoaugment}
Ekin~D Cubuk, Barret Zoph, Dandelion Mane, Vijay Vasudevan, and Quoc~V Le.
\newblock Autoaugment: Learning augmentation policies from data.
\newblock {\em arXiv preprint arXiv:1805.09501}, 2018.

\bibitem{imagenet2009}
Jia Deng, Wei Dong, Richard Socher, Li-Jia Li, Kai Li, and Li Fei-Fei.
\newblock Imagenet: A large-scale hierarchical image database.
\newblock In {\em Proceedings of IEEE Conference on Computer Vision and Pattern
  Recognition (CVPR)}, 2009.

\bibitem{devries2017dataset}
Terrance DeVries and Graham~W Taylor.
\newblock Dataset augmentation in feature space.
\newblock {\em arXiv preprint arXiv:1702.05538}, 2017.

\bibitem{cutout2017}
Terrance DeVries and Graham~W Taylor.
\newblock Improved regularization of convolutional neural networks with cutout.
\newblock {\em arXiv preprint arXiv:1708.04552}, 2017.

\bibitem{dwibedi2017cut}
Debidatta Dwibedi, Ishan Misra, and Martial Hebert.
\newblock Cut, paste and learn: Surprisingly easy synthesis for instance
  detection.
\newblock In {\em Proceedings of the IEEE International Conference on Computer
  Vision}, pages 1301--1310, 2017.

\bibitem{fang2019instaboost}
Hao-Shu Fang, Jianhua Sun, Runzhong Wang, Minghao Gou, Yong-Lu Li, and Cewu Lu.
\newblock Instaboost: Boosting instance segmentation via probability map guided
  copy-pasting.
\newblock {\em arXiv preprint arXiv:1908.07801}, 2019.

\bibitem{ford2019adversarial}
Nic Ford, Justin Gilmer, Nicolas Carlini, and Dogus Cubuk.
\newblock Adversarial examples are a natural consequence of test error in
  noise.
\newblock {\em arXiv preprint arXiv:1901.10513}, 2019.

\bibitem{gastaldi2017shake}
Xavier Gastaldi.
\newblock Shake-shake regularization.
\newblock {\em arXiv preprint arXiv:1705.07485}, 2017.

\bibitem{girshick2018detectron}
Ross Girshick, Ilija Radosavovic, Georgia Gkioxari, Piotr Doll{\'a}r, and
  Kaiming He.
\newblock Detectron, 2018.

\bibitem{golovin2017google}
Daniel Golovin, Benjamin Solnik, Subhodeep Moitra, Greg Kochanski, John Karro,
  and D Sculley.
\newblock Google vizier: A service for black-box optimization.
\newblock In {\em Proceedings of the 23rd ACM SIGKDD International Conference
  on Knowledge Discovery and Data Mining}, pages 1487--1495. ACM, 2017.

\bibitem{han2017deep}
Dongyoon Han, Jiwhan Kim, and Junmo Kim.
\newblock Deep pyramidal residual networks.
\newblock In {\em Proceedings of IEEE Conference on Computer Vision and Pattern
  Recognition (CVPR)}, pages 6307--6315. IEEE, 2017.

\bibitem{hannun2014deep}
Awni Hannun, Carl Case, Jared Casper, Bryan Catanzaro, Greg Diamos, Erich
  Elsen, Ryan Prenger, Sanjeev Satheesh, Shubho Sengupta, Adam Coates, et~al.
\newblock Deep speech: Scaling up end-to-end speech recognition.
\newblock {\em arXiv preprint arXiv:1412.5567}, 2014.

\bibitem{he2016deep}
Kaiming He, Xiangyu Zhang, Shaoqing Ren, and Jian Sun.
\newblock Deep residual learning for image recognition.
\newblock In {\em Proceedings of the IEEE Conference on Computer Vision and
  Pattern Recognition (CVPR)}, pages 770--778, 2016.

\bibitem{hershey2017cnn}
Shawn Hershey, Sourish Chaudhuri, Daniel~PW Ellis, Jort~F Gemmeke, Aren Jansen,
  R~Channing Moore, Manoj Plakal, Devin Platt, Rif~A Saurous, Bryan Seybold,
  et~al.
\newblock Cnn architectures for large-scale audio classification.
\newblock In {\em 2017 ieee international conference on acoustics, speech and
  signal processing (icassp)}, pages 131--135. IEEE, 2017.

\bibitem{hinton2012deep}
Geoffrey Hinton, Li Deng, Dong Yu, George Dahl, Abdel-rahman Mohamed, Navdeep
  Jaitly, Andrew Senior, Vincent Vanhoucke, Patrick Nguyen, Brian Kingsbury,
  et~al.
\newblock Deep neural networks for acoustic modeling in speech recognition.
\newblock {\em IEEE Signal processing magazine}, 29, 2012.

\bibitem{ho2019population}
Daniel Ho, Eric Liang, Ion Stoica, Pieter Abbeel, and Xi Chen.
\newblock Population based augmentation: Efficient learning of augmentation
  policy schedules.
\newblock {\em arXiv preprint arXiv:1905.05393}, 2019.

\bibitem{kanda2013elastic}
Naoyuki Kanda, Ryu Takeda, and Yasunari Obuchi.
\newblock Elastic spectral distortion for low resource speech recognition with
  deep neural networks.
\newblock In {\em 2013 IEEE Workshop on Automatic Speech Recognition and
  Understanding}, pages 309--314. IEEE, 2013.

\bibitem{krizhevsky2009learning}
Alex Krizhevsky and Geoffrey Hinton.
\newblock Learning multiple layers of features from tiny images.
\newblock Technical report, University of Toronto, 2009.

\bibitem{krizhevsky2012imagenet}
Alex Krizhevsky, Ilya Sutskever, and Geoffrey~E. Hinton.
\newblock Imagenet classification with deep convolutional neural networks.
\newblock In {\em Advances in Neural Information Processing Systems}, 2012.

\bibitem{lemley2017smart}
Joseph Lemley, Shabab Bazrafkan, and Peter Corcoran.
\newblock Smart augmentation learning an optimal data augmentation strategy.
\newblock {\em IEEE Access}, 5:5858--5869, 2017.

\bibitem{lim2019fast}
Sungbin Lim, Ildoo Kim, Taesup Kim, Chiheon Kim, and Sungwoong Kim.
\newblock Fast autoaugment.
\newblock {\em arXiv preprint arXiv:1905.00397}, 2019.

\bibitem{lin2017focal}
Tsung-Yi Lin, Priya Goyal, Ross Girshick, Kaiming He, and Piotr Doll{\'a}r.
\newblock Focal loss for dense object detection.
\newblock In {\em Proceedings of the IEEE international conference on computer
  vision}, pages 2980--2988, 2017.

\bibitem{lin2014microsoft}
Tsung-Yi Lin, Michael Maire, Serge Belongie, James Hays, Pietro Perona, Deva
  Ramanan, Piotr Doll{\'a}r, and C~Lawrence Zitnick.
\newblock Microsoft coco: Common objects in context.
\newblock In {\em European conference on computer vision}, pages 740--755.
  Springer, 2014.

\bibitem{liu2017progressive}
Chenxi Liu, Barret Zoph, Jonathon Shlens, Wei Hua, Li-Jia Li, Li Fei-Fei, Alan
  Yuille, Jonathan Huang, and Kevin Murphy.
\newblock Progressive neural architecture search.
\newblock {\em arXiv preprint arXiv:1712.00559}, 2017.

\bibitem{liu2017hierarchical}
Hanxiao Liu, Karen Simonyan, Oriol Vinyals, Chrisantha Fernando, and Koray
  Kavukcuoglu.
\newblock Hierarchical representations for efficient architecture search.
\newblock In {\em International Conference on Learning Representations}, 2018.

\bibitem{liu2018darts}
Hanxiao Liu, Karen Simonyan, and Yiming Yang.
\newblock Darts: Differentiable architecture search.
\newblock {\em arXiv preprint arXiv:1806.09055}, 2018.

\bibitem{liu2016ssd}
Wei Liu, Dragomir Anguelov, Dumitru Erhan, Christian Szegedy, Scott Reed,
  Cheng-Yang Fu, and Alexander~C Berg.
\newblock Ssd: Single shot multibox detector.
\newblock In {\em European conference on computer vision}, pages 21--37.
  Springer, 2016.

\bibitem{lopes2019improving}
Raphael~Gontijo Lopes, Dong Yin, Ben Poole, Justin Gilmer, and Ekin~D Cubuk.
\newblock Improving robustness without sacrificing accuracy with patch gaussian
  augmentation.
\newblock {\em arXiv preprint arXiv:1906.02611}, 2019.

\bibitem{mun2017generative}
Seongkyu Mun, Sangwook Park, David~K Han, and Hanseok Ko.
\newblock Generative adversarial network based acoustic scene training set
  augmentation and selection using svm hyper-plane.
\newblock In {\em Detection and Classification of Acoustic Scenes and Events
  Workshop}, 2017.

\bibitem{netzer2011reading}
Yuval Netzer, Tao Wang, Adam Coates, Alessandro Bissacco, Bo Wu, and Andrew~Y
  Ng.
\newblock Reading digits in natural images with unsupervised feature learning.
\newblock In {\em NIPS Workshop on Deep Learning and Unsupervised Feature
  Learning}, 2011.

\bibitem{ngiam2019starnet}
Jiquan Ngiam, Benjamin Caine, Wei Han, Brandon Yang, Yuning Chai, Pei Sun, Yin
  Zhou, Xi Yi, Ouais Alsharif, Patrick Nguyen, et~al.
\newblock Starnet: Targeted computation for object detection in point clouds.
\newblock {\em arXiv preprint arXiv:1908.11069}, 2019.

\bibitem{park2019specaugment}
Daniel~S Park, William Chan, Yu Zhang, Chung-Cheng Chiu, Barret Zoph, Ekin~D
  Cubuk, and Quoc~V Le.
\newblock Specaugment: A simple data augmentation method for automatic speech
  recognition.
\newblock {\em arXiv preprint arXiv:1904.08779}, 2019.

\bibitem{perez2017effectiveness}
Luis Perez and Jason Wang.
\newblock The effectiveness of data augmentation in image classification using
  deep learning.
\newblock {\em arXiv preprint arXiv:1712.04621}, 2017.

\bibitem{pham2018efficient}
Hieu Pham, Melody~Y Guan, Barret Zoph, Quoc~V Le, and Jeff Dean.
\newblock Efficient neural architecture search via parameter sharing.
\newblock In {\em International Conference on Machine Learning}, 2018.

\bibitem{ratner2017learning}
Alexander~J Ratner, Henry Ehrenberg, Zeshan Hussain, Jared Dunnmon, and
  Christopher R{\'e}.
\newblock Learning to compose domain-specific transformations for data
  augmentation.
\newblock In {\em Advances in Neural Information Processing Systems}, pages
  3239--3249, 2017.

\bibitem{ravuri2019classification}
Suman Ravuri and Oriol Vinyals.
\newblock Classification accuracy score for conditional generative models.
\newblock {\em arXiv preprint arXiv:1905.10887}, 2019.

\bibitem{recht2018cifar}
Benjamin Recht, Rebecca Roelofs, Ludwig Schmidt, and Vaishaal Shankar.
\newblock Do imagenet classifiers generalize to imagenet?
\newblock {\em arXiv preprint arXiv:1902.10811}, 2019.

\bibitem{sato2015apac}
Ikuro Sato, Hiroki Nishimura, and Kensuke Yokoi.
\newblock Apac: Augmented pattern classification with neural networks.
\newblock {\em arXiv preprint arXiv:1505.03229}, 2015.

\bibitem{simard2003best}
Patrice~Y Simard, David Steinkraus, John~C Platt, et~al.
\newblock Best practices for convolutional neural networks applied to visual
  document analysis.
\newblock In {\em Proceedings of International Conference on Document Analysis
  and Recognition}, 2003.

\bibitem{sixt2016rendergan}
Leon Sixt, Benjamin Wild, and Tim Landgraf.
\newblock Rendergan: Generating realistic labeled data.
\newblock {\em arXiv preprint arXiv:1611.01331}, 2016.

\bibitem{snoek2012practical}
Jasper Snoek, Hugo Larochelle, and Ryan~P Adams.
\newblock Practical bayesian optimization of machine learning algorithms.
\newblock In {\em Advances in neural information processing systems}, pages
  2951--2959, 2012.

\bibitem{szegedy2013intriguing}
Christian Szegedy, Wojciech Zaremba, Ilya Sutskever, Joan Bruna, Dumitru Erhan,
  Ian Goodfellow, and Rob Fergus.
\newblock Intriguing properties of neural networks.
\newblock {\em arXiv preprint arXiv:1312.6199}, 2013.

\bibitem{tan2019efficientnet}
Mingxing Tan and Quoc~V Le.
\newblock Efficientnet: Rethinking model scaling for convolutional neural
  networks.
\newblock {\em arXiv preprint arXiv:1905.11946}, 2019.

\bibitem{tran2017bayesian}
Toan Tran, Trung Pham, Gustavo Carneiro, Lyle Palmer, and Ian Reid.
\newblock A bayesian data augmentation approach for learning deep models.
\newblock In {\em Advances in Neural Information Processing Systems}, pages
  2794--2803, 2017.

\bibitem{wan2013regularization}
Li Wan, Matthew Zeiler, Sixin Zhang, Yann Le~Cun, and Rob Fergus.
\newblock Regularization of neural networks using dropconnect.
\newblock In {\em International Conference on Machine Learning}, pages
  1058--1066, 2013.

\bibitem{xie2019unsupervised}
Qizhe Xie, Zihang Dai, Eduard Hovy, Minh-Thang Luong, and Quoc~V Le.
\newblock Unsupervised data augmentation.
\newblock {\em arXiv preprint arXiv:1904.12848}, 2019.

\bibitem{yamada2018shakedrop}
Yoshihiro Yamada, Masakazu Iwamura, and Koichi Kise.
\newblock Shakedrop regularization.
\newblock {\em arXiv preprint arXiv:1802.02375}, 2018.

\bibitem{yin2019afourier}
Dong Yin, Raphael~Gontijo Lopes, Jonathon Shlens, Ekin~D Cubuk, and Justin
  Gilmer.
\newblock A fourier perspective on model robustness in computer vision.
\newblock {\em arXiv preprint arXiv:1906.08988}, 2019.

\bibitem{WRN2016}
Sergey Zagoruyko and Nikos Komodakis.
\newblock Wide residual networks.
\newblock In {\em British Machine Vision Conference}, 2016.

\bibitem{zhang2017mixup}
Hongyi Zhang, Moustapha Cisse, Yann~N Dauphin, and David Lopez-Paz.
\newblock mixup: Beyond empirical risk minimization.
\newblock {\em arXiv preprint arXiv:1710.09412}, 2017.

\bibitem{zhong2017random}
Zhun Zhong, Liang Zheng, Guoliang Kang, Shaozi Li, and Yi Yang.
\newblock Random erasing data augmentation.
\newblock {\em arXiv preprint arXiv:1708.04896}, 2017.

\bibitem{zhu2017data}
Xinyue Zhu, Yifan Liu, Zengchang Qin, and Jiahong Li.
\newblock Data augmentation in emotion classification using generative
  adversarial networks.
\newblock {\em arXiv preprint arXiv:1711.00648}, 2017.

\bibitem{zoph2019learning}
Barret Zoph, Ekin~D Cubuk, Golnaz Ghiasi, Tsung-Yi Lin, Jonathon Shlens, and
  Quoc~V Le.
\newblock Learning data augmentation strategies for object detection.
\newblock {\em arXiv preprint arXiv:1906.11172}, 2019.

\bibitem{zoph2016neural}
Barret Zoph and Quoc~V Le.
\newblock Neural architecture search with reinforcement learning.
\newblock In {\em International Conference on Learning Representations}, 2017.

\bibitem{zoph2017learning}
Barret Zoph, Vijay Vasudevan, Jonathon Shlens, and Quoc~V Le.
\newblock Learning transferable architectures for scalable image recognition.
\newblock In {\em Proceedings of IEEE Conference on Computer Vision and Pattern
  Recognition}, 2017.

\end{thebibliography}
